\documentclass[conference]{IEEEtran}
\IEEEoverridecommandlockouts
\usepackage{cite}
\usepackage{amsmath,amssymb,amsfonts}
\usepackage{algorithmic}
\usepackage{graphicx}
\usepackage{textcomp}
\usepackage{xcolor}
\usepackage{booktabs}
\def\BibTeX{{\rm B\kern-.05em{\sc i\kern-.025em b}\kern-.08em
    T\kern-.1667em\lower.7ex\hbox{E}\kern-.125emX}}
\begin{document}

\title{Friends Across Time: Multi-Scale Action Segmentation Transformer for Surgical Phase Recognition \\
}

\author{\IEEEauthorblockN{1\textsuperscript{st} Bokai Zhang}
\IEEEauthorblockA{\textit{Blue River Technology} \\
California, United States of America \\
zhangbokai1994@gmail.com}
\and
\IEEEauthorblockN{2\textsuperscript{nd} Jiayuan Meng}
\IEEEauthorblockA{\textit{Blue River Technology} \\
California, United States of America \\
meng.jiayuan@gmail.com}
\and
\IEEEauthorblockN{3\textsuperscript{rd} Bin Cheng}
\IEEEauthorblockA{\textit{Blue River Technology} \\
California, United States of America \\
cb3974@winlab.rutgers.edu}
\and
\IEEEauthorblockN{4\textsuperscript{th} Dean Biskup}
\IEEEauthorblockA{\textit{Blue River Technology} \\
California, United States of America \\
deanbiskup@gmail.com}
\and
\IEEEauthorblockN{5\textsuperscript{th} Svetlana Petculescu }
\IEEEauthorblockA{\textit{Independent Consultant} \\
Washington, United States of America \\
sveta0704@gmail.com}
\and
\IEEEauthorblockN{6\textsuperscript{th} Angela Chapman}
\IEEEauthorblockA{\textit{Blue Grotto Labs, LLC} \\
Massachusetts, United States of America \\
achapman@bluegrottolabs.com}
}

\maketitle

\begin{abstract}
Automatic surgical phase recognition is a core technology for modern operating rooms and online surgical video assessment platforms. Current state-of-the-art methods use both spatial and temporal information to tackle the surgical phase recognition task. Building on this idea, we propose the Multi-Scale Action Segmentation Transformer (MS-AST) for offline surgical phase recognition and the Multi-Scale Action Segmentation Causal Transformer (MS-ASCT) for online surgical phase recognition. We use ResNet50 or EfficientNetV2-M for spatial feature extraction. Our MS-AST and MS-ASCT can model temporal information at different scales with multi-scale temporal self-attention and multi-scale temporal cross-attention,  which enhances the capture of temporal relationships between frames and segments. We demonstrate that our method can achieve 95.26\% and 96.15\% accuracy on the Cholec80 dataset for online and offline surgical phase recognition, respectively, which achieves new state-of-the-art results. Our method can also achieve state-of-the-art results on non-medical datasets in the video action segmentation domain. 
\end{abstract}

\begin{IEEEkeywords}
surgical phase recognition, multi-scale, action segmentation, transformer
\end{IEEEkeywords}

\section{Introduction}
Over the last few years, Video-Based Analysis (VBA) has been increasingly used for surgical video analysis~\cite{feldman2020sages}.  
One of its key technologies, surgical phase recognition, is instrumental in facilitating efficient skill assessment training. Surgical phase recognition is the automatic detection of start and stop times for different steps or phases of a surgery. While online surgical phase recognition algorithms are mainly designed to support surgeons in the operating room (OR), 
offline surgical phase recognition algorithms play a crucial role in efficiently categorizing large collections of surgical videos. This efficient categorization streamlines training by allowing for the comparison of surgical phase timings between different surgeons, aiding in skill assessment. Moreover, it supports the analysis of surgical phase sequences, contributing to the standardization of surgical techniques 
. It also helps in quickly identifying key moments in surgery videos for expert analysis, significantly reducing time in both training and evaluation processes.

To capture spatial information from video frames, early research used image classification networks \cite{twinanda2016endonet}. However, these methods did not capture temporal information between video frames. Researchers then combined image classification networks with Long Short-Term Memory (LSTM) networks for both spatial and temporal modeling \cite{jin2017sv,jin2020multi}. Recent research~\cite{czempiel2020tecno,farha2019ms} utilized Multi-Stage Temporal Convolutional Networks (MS-TCN) for full video temporal modeling and has demonstrated superior performance compared to LSTM~\cite{czempiel2020tecno,zhang2021swnet}. As vision transformers increasingly dominate many computer vision research areas \cite{dosovitskiy2020image}, researchers now utilize transformers for temporal modeling \cite{yi2021asformer,czempiel2021opera, zhang2022towards,jin2022trans, chen2022spatio, zhang2023sf, zhang2023surgical}. Researchers believe that the attention layers design in transformers can force the model to learn and focus on informative timestamps in surgical videos instead of timestamps like motion blur, and idle action which are not informative for surgical phase recognition.

In this paper, we expand upon transformer-based techniques and propose the Multi-Scale Action Segmentation Transformer (MS-AST) for surgical phase recognition. With different temporal modeling scales, our method can achieve frame-level modeling and segment-level modeling at the same time. Our model is capable of capturing both fast and slow actions, such as short and long surgical phases, respectively, across a range of small and large temporal scales. Our contributions in this paper are listed as follows: (1) Building upon Transformer for Action Segmentation (ASFormer)~\cite{yi2021asformer}, we design multi-scale temporal self-attention and multi-scale temporal cross-attention to capture frame and segment relations at different temporal scales in our proposed MS-AST for surgical phase recognition. (2) We modify MS-AST to Multi-Scale Action Segmentation Causal Transformer (MS-ASCT), a causal design for online surgical phase recognition. (3) Our proposed MS-AST and MS-ASCT achieve new state-of-the-art results in offline and online surgical phase recognition, respectively. (4) To demonstrate the robustness and wide applicability of our method, we evaluate our MS-AST on the 50Salads \cite{stein2013combining} and GTEA \cite{fathi2011learning} datasets, which are non-medical datasets that are widely used in the video action segmentation domain. Our MS-AST also achieves state-of-the-art results on these datasets.


\section{Method}

\begin{figure}[h]%
\centering
\includegraphics[width=0.47\textwidth]{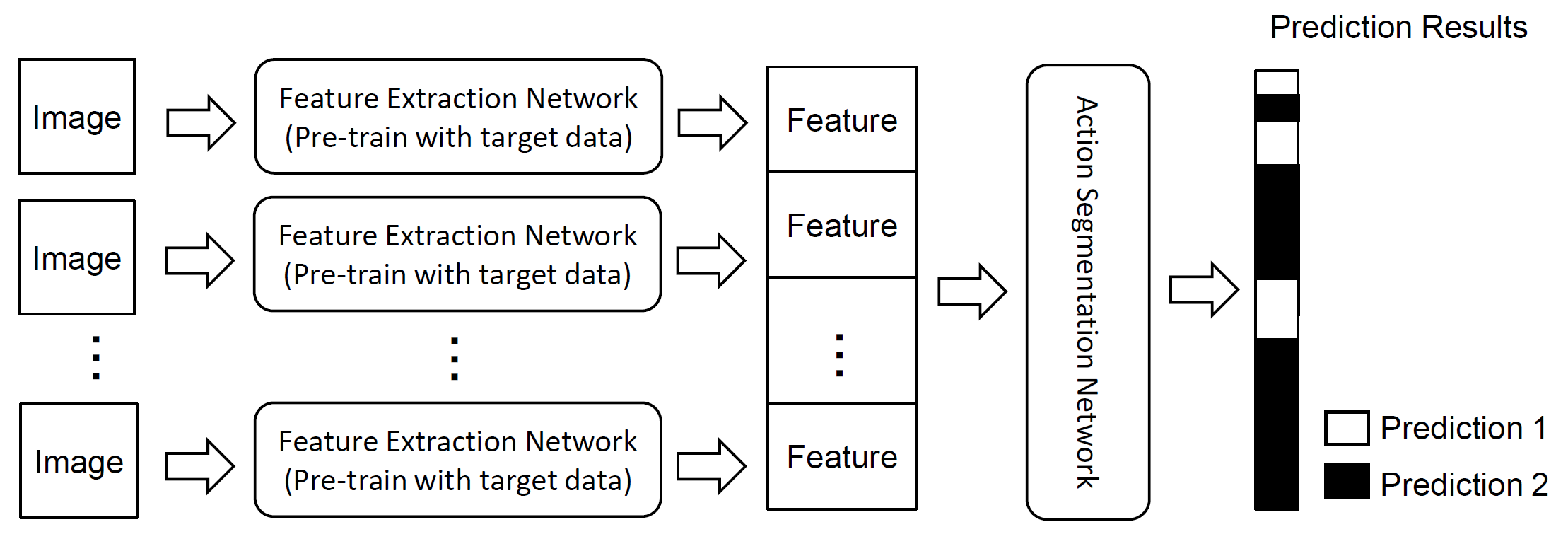}
\caption{The overview of our method }\label{fig1}
\end{figure}

An overview of our method is depicted in Figure~\ref{fig1}. First, we train an image classification network with training data from the target dataset for feature extraction. Second, we use the image classification network to extract frame-level spatial information for each second of the video frame. We concatenate these frame-level features to get full video features. Finally, we train our action segmentation network with full video features for temporal modeling and surgical phase recognition. These steps are described in more detail below.

\subsection{Feature Extraction Network}

Full video modeling with all video frames from surgical videos requires significant GPU resources. Feature extraction networks are used to tackle this problem. Instead of learning from raw video frames, with feature extraction networks, video frames can be summarized into feature vectors which are more memory efficient during training.

We utilize ResNet50 \cite{he2016deep}, which is widely used in previous studies \cite{czempiel2020tecno,jin2022trans}. We load up the ImageNet pre-trained weights and conduct transfer learning with the surgical phase dataset. Our Multi-Scale Action Segmentation Transformer can also work with features generated by different feature extraction networks. We utilize EfficientNetV2-M (EffNetV2) \cite{tan2021efficientnetv2} as another feature extraction backbone to validate our design.

\subsection{Action Segmentation Network}

\subsubsection{Transformer for Action Segmentation}
ASFormer \cite{yi2021asformer} is an encoder-decoder structured transformer proposed to tackle action segmentation tasks.  With extracted frame-wise feature sequences, the encoder will first predict the initial action probability for each frame. Then the initial predictions will be passed to multiple decoders for incremental refinement. Each encoder comprises encoder blocks that have a feed-forward and a single-head self-attention layer. Each decoder comprises decoder blocks that have a feed-forward and a single-head cross-attention layer. To model local features first and then gradually capture the global information with enlarged receptive fields, dilated convolutions were used in the feed-forward layers with a gradually increasing dilation rate, and sliding window attention was used with increasing window size in the attention layers.

\subsubsection{Multi-Scale Action Segmentation Transformer}
To capture segment-level temporal information and model fast action and slow action with different temporal scales, we modify the self-attention layer in each encoder block and the cross-attention layer in each decoder block in ASFormer to multi-scale temporal self-attention and multi-scale temporal cross-attention. 

\begin{figure}[h]%
\centering
\includegraphics[width=0.44\textwidth]{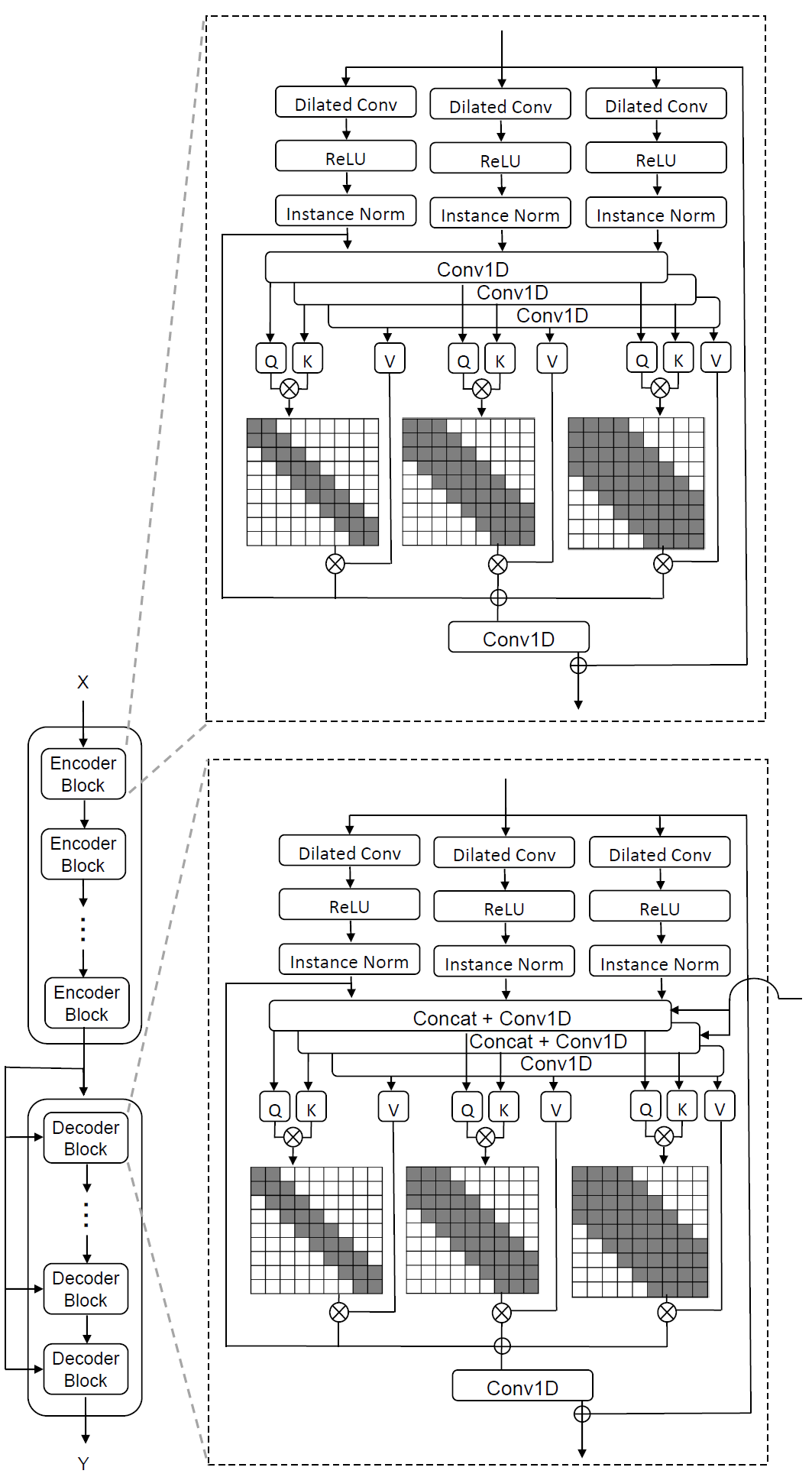}
\caption{Multi-Scale Action Segmentation Transformer}\label{fig2}
\end{figure}

As shown in Figure \ref{fig2}, we use three different temporal scales in our multi-scale temporal self-attention. The dilated convolution kernel sizes are 3, 5, and 17 in the first layers for each encoder and decoder, corresponding to 3 different temporal scales. The window sizes for the sliding window attention also change depending on the kernel sizes used in the dilated convolution. Following the design of the ASFormer, when the kernel size is 3, the window size doubles in each layer, starting from 1 in the first layer and ending at 512 in the 10th layer. In our design, when the kernel size is 5 and 17, the window size at the first layer is 1. For a kernel size of 5, the window size then doubles in each layer, starting from 4 in the second layer and ending at 1024 in the 10th layer. For a kernel size of 17, the window size doubles starting from 16 in the second layer and ending at 4096 in the 10th layer. The output features with attention in different scales can be calculated by
\begin{equation}
\begin{aligned}
out_{i} & = \alpha \times w_{1,i} \times Attention_{1, i}(out_{i})\\
  & + \alpha \times w_{2,i} \times Attention_{2, i}(out_{i})\\
  & + \alpha \times w_{3,i} \times Attention_{3, i}(out_{i}) + out_{i} 
\end{aligned}
\end{equation}

Where $out_{i}$ represents the output features generated in the $i$th encoder or decoder block. $w_{1,i}$, $w_{2,i}$, $w_{3,i}$ are weighted parameters that are learned during the training. $Attention_{1, i}$, $Attention_{2, i}$, and $Attention_{3, i}$ represent attention results at 3 different scales. $\alpha$ is equal to 1 in the first decoder and then is exponentially decreased for the remaining decoders. 


In contrast to the multi-scale self-attention in the encoder block, in the decoder block's multi-scale cross-attention layer, the query $Q$ and key $K$ are obtained from the output of the encoder and the output of the previous layer, while the value $V$ is obtained from the output of the previous layer, as shown in Figure \ref{fig2}.

\subsubsection{Multi-Scale Action Segmentation Causal Transformer}

\begin{figure}[h]%
\centering
\includegraphics[width=0.38\textwidth]{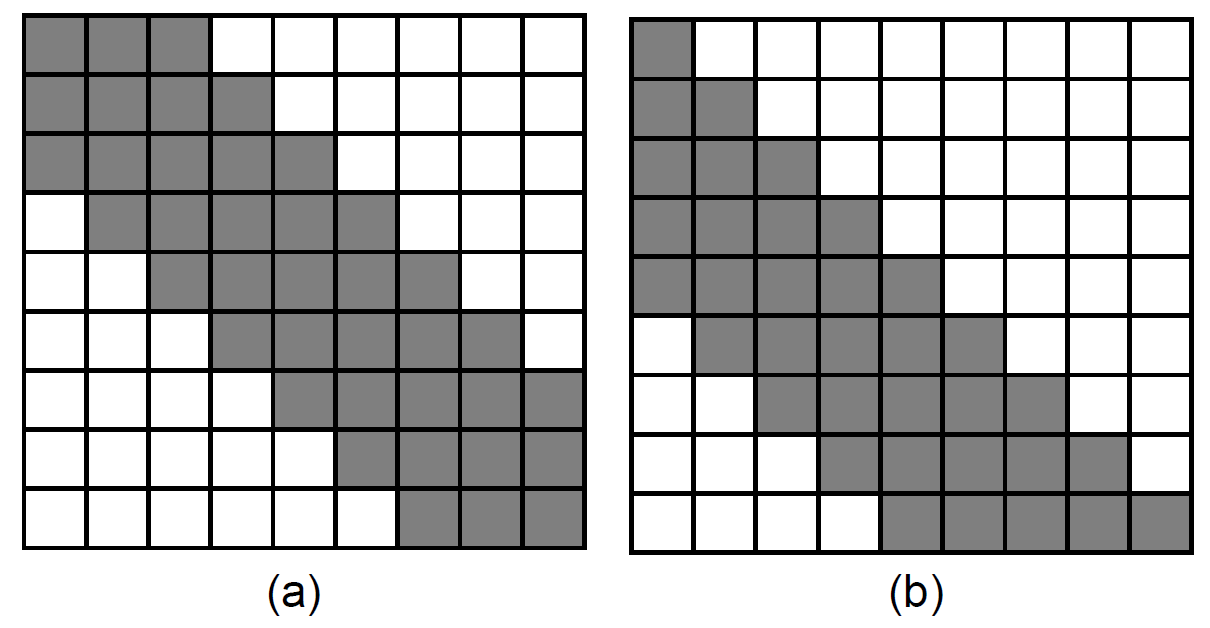}
\caption{Sliding Window Attention: (1) Non-causal (2) Causal}\label{fig3xx}
\end{figure}

In order to achieve online surgical phase recognition in the OR, we modify our MS-AST to incorporate causality, creating the Multi-Scale Action Segmentation Causal Transformer (MS-ASCT). First, we modify the dilated convolution to causal dilated convolution following previous TeCNO research \cite{czempiel2020tecno}. Second, we remove layer normalization from our network to avoid future information leaks. Finally, we use the causal design of sliding window attention shown in Figure \ref{fig3xx}, so that the attention operation only uses past information. We use window size equals 5 in Figure \ref{fig3xx} as an example.

\section{Dataset}
The Cholec80 dataset \cite{twinanda2016endonet} consists of 80 videos of cholecystectomy surgeries performed by 13 surgeons. The dataset includes annotations for both surgical phase and tool presence. The 7 surgical phases, labeled as P1 to P7, are “Preparation”, “Calot triangle dissection”, “Clipping and cutting”, “Gallbladder dissection”, “Gallbladder packaging”, “Cleaning and coagulation”, and “Gallbladder retraction”. Following previous research \cite{jin2017sv,jin2020multi,jin2021temporal,jin2022trans,he2022empirical,zhang2023c}, we use the first 40 videos for training and reserve the remaining 40 for testing. We conduct both online surgical phase recognition and offline surgical phase recognition experiments with the Cholec80 dataset.

To evaluate our method in the action segmentation domain, the 50Salads dataset \cite{stein2013combining} and the GTEA dataset \cite{fathi2011learning} are used following the MS-TCN \cite{farha2019ms} and ASFormer \cite{yi2021asformer} studies. 
The 50Salads dataset contains 50 videos of salad preparation steps, with 17 action classes and another two classes for "action start" and "action end". 
The GTEA dataset contains 28 videos of 11 classes of daily activities in a kitchen. 
For fair comparisons, we utilized features extracted in \cite{farha2019ms}. These features are extracted with Inflated 3D ConvNet (I3D) \cite{carreira2017quo}. 
Following previous research \cite{farha2019ms,yi2021asformer}, five-fold cross-validations are performed on the 50Salads dataset, and four-fold cross-validations are performed on the GTEA dataset with provided data splits.


\section{Experiments}
We utilized PyTorch to implement all experiments. We performed all training and testing on a single desktop with one Intel Core i9-13900K CPU and one NVIDIA GeForce RTX 4090 GPU.
\subsection{Evaluation metrics}
We use frame-level metrics and segmental metrics \cite{lea2016learning,lea2017temporal} to evaluate model performance. Following previous research \cite{jin2017sv,jin2020multi,jin2022trans,he2022empirical}, we compute precision, recall, and Jaccard scores for each surgical phase, then average over all surgical phases. We use segmental edit distance score and segmental F1-score as segmental metrics to evaluate over-segmentation error. Absolute improvements are reported in the following sections. 
The overlap can be measured with Intersection over Union (IOU).
For a simpler comparison, we also calculate $\text{F1}_{\mathit{AVG}}$ which is the average of the segmental F1-score at $10\%$, $25\%$, and $50\%$ overlap thresholds, as follows:

\begin{equation}
    \text{F1}_{\mathit{AVG}}=\dfrac{1}{3} \times (\text{F1@10 + F1@25 + F1@50}) 
\end{equation}

\subsection{Implementation details}
We train ResNet50 and EffNetV2, two feature extraction networks, using cross-entropy loss with a learning rate of 1e-4 and weight decay of 1e-5. We set the batch size to 16 and the training epochs to 50. To augment the data, we resized the smaller side of the frames to 256 pixels and randomly cropped 224 by 224 patches as training samples for ResNet50. Similarly, we resized the smaller side of the frames to 400 pixels and randomly cropped 384 by 384 patches as training samples for EffNetV2. We also selected 15\% of the training samples randomly and rotated them within 10 degrees to simulate camera rotation.

We trained our MS-AST and MS-ASCT with cross-entropy loss and smooth loss \cite{farha2019ms}. We utilized the Adam optimizer with a learning rate of 1e-4, batch size of 1, dropout rate of 0.5, and 200 training epochs. We used one encoder and three decoders in MS-AST for offline surgical phase recognition, and we used one encoder and one decoder in MS-ASCT for online surgical phase recognition. We set the total number of the dilated convolution layers in each encoder and each decoder to 10, and set the number of feature maps to 64.

\subsection{Results}

\subsubsection{Online surgical phase recognition}
To compare model performance that used different temporal scales, we compare EffNetV2 Causal ASFormer \cite{zhang2023c}, which utilizes one temporal scale with a kernel size of 3, and EffNetV2 MS-ASCT, which uses multiple temporal scales with different kernel sizes, in Table~\ref{tab0}. Our kernel size is selected following $1 + 2^{n}$, where $n$ is a positive integer. As shown in Table~\ref{tab0},  MS-ASCT with 3 temporal scales outperforms Causal ASFormer \cite{zhang2023c} with 1 temporal scale and MS-ASCT with 2 temporal scales. MS-ASCT with kernel sizes of 3, 5, and 17 slightly outperforms MS-ASCT with kernel sizes of 3, 5, and 9.

\begin{table}[h]
\begin{center}
\caption{Overall accuracy (\%), segmental edit distance score, segmental F1-score at different thresholds, and the average of segmental F1-score at different thresholds \\}\label{tab0}
\resizebox{\linewidth}{!}{\begin{tabular}{llllllll}
\toprule
Method name & Kernel & ACC & Edit & F1@10 & F1@25  & F1@50 & F1@AVG\\
\midrule
EffNetV2 Causal ASFormer \cite{zhang2023c}  &3& 93.44 & 46.20 & 55.43  & 54.60 & 50.47   & 53.50\\
\midrule
EffNetV2 MS-ASCT  &3,5 & 93.94 & 50.85 & 58.26 & 57.83 & 53.48 & 56.52 \\ 
EffNetV2 MS-ASCT  &3,5,9 & 94.28 & 58.25 & 66.18 & 65.20 & 60.78 & 64.05 \\ 
EffNetV2 MS-ASCT  &3,5,17 & \textbf{94.64} & \textbf{61.82} & \textbf{68.02} & \textbf{68.02} & \textbf{62.45} & \textbf{66.16}\\  
\bottomrule
\end{tabular}}
\end{center}
\end{table}

To demonstrate that our methods outperform other state-of-the-art approaches, we compare our ResNet MS-ASCT and EffNetV2 MS-ASCT with PhaseLSTM \cite{twinanda2016endonet}, EndoLSTM \cite{twinanda2016endonet}, MTRCNet \cite{jin2020multi}, SV-RCNet \cite{jin2017sv}, TeCNO \cite{czempiel2020tecno}, SlowFast UniGRU, TimeSformer UniGRU, Swin UniGRU \cite{he2022empirical}, Trans-SVNet \cite{jin2022trans}, PATG \cite{kadkhodamohammadi2022patg}, and Spatio-Temporal Causal Transformer(denoted as ST-CT) \cite{chen2022spatio} in Table~\ref{tab2}. The mean and the standard deviation in some studies in Table~\ref{tab2} refer to results over repeated experimental runs. Our ResNet MS-ASCT outperforms all other methods from previous studies. Our ResNet MS-ASCT outperforms previous state-of-the-art ST-CT \cite{chen2022spatio} by approximately 2\% in accuracy and recall, as well as approximately 3\% in precision. Our EffNetV2 MS-ASCT also outperforms ResNet MS-ASCT, which shows that our designed method can work with features extracted by different feature extraction networks. Our EffNetV2 MS-ASCT outperforms previous state-of-the-art ST-CT \cite{chen2022spatio} by approximately 3\% in accuracy, approximately 5\% in precision, and approximately 4\% in recall.

\begin{table}[h]
\centering
\caption{Overall accuracy, precision, and recall for different methods on Cholec80 dataset for online surgical phase recognition (mean $\pm$ std. \%)}\label{tab2}
\begin{tabular}{llll}
\hline
Method Name &  Accuracy & Precision & Recall\\
\hline

PhaseLSTM \cite{twinanda2016endonet,he2022empirical}  & 79.68$\pm$0.07 & 72.85$\pm$0.10 & 73.45$\pm$0.12 \\
EndoLSTM \cite{twinanda2016endonet,he2022empirical} & 80.85$\pm$0.17  & 76.81$\pm$2.62 & 72.07$\pm$0.64 \\
MTRCNet \cite{jin2020multi,he2022empirical } & 82.76$\pm$0.01 & 76.08$\pm$0.01 & 78.02$\pm$0.13 \\
SV-RCNet \cite{jin2017sv,he2022empirical} & 86.58$\pm$1.01 & 80.53$\pm$1.59 & 79.94$\pm$1.79 \\
TeCNO \cite{czempiel2020tecno,he2022empirical} & 88.56$\pm$0.27 & 81.64$\pm$0.41 & 85.24$\pm$1.06 \\
SlowFast UniGRU \cite{he2022empirical} & 90.47$\pm$0.46 & 83.12$\pm$2.09 & 82.33$\pm$1.22 \\
TimeSformer UniGRU \cite{he2022empirical} & 90.42$\pm$0.47  & 86.05$\pm$1.13 & 83.20$\pm$1.80 \\
Swin UniGRU \cite{he2022empirical} & 90.88$\pm$0.01 & 85.07$\pm$1.74 & 85.59$\pm$0.53  \\
\hline
Trans-SVNet \cite{jin2022trans,chen2022spatio} & 89.6 & 81.7 & 87.5  \\
PATG \cite{kadkhodamohammadi2022patg} & 91.36 & 86.88 & 84.00  \\
ST-CT \cite{chen2022spatio} & 91.4 & 85.4 & 86.3  \\
\hline
ResNet MS-ASCT & 93.58$\pm$0.13 & 88.90$\pm$0.41 & 88.20$\pm$0.34  \\
EffNetV2 MS-ASCT & \textbf{94.59$\pm$0.04} & \textbf{90.41$\pm$0.22} & \textbf{90.07$\pm$0.46}  \\

\hline
\end{tabular}
\end{table}

We also calculated video-by-video accuracy, precision, recall, and Jaccard score for our EffNetV2 MS-ASCT and compared them with other state-of-the-art approaches as shown in Table~\ref{table1x}. The mean and the standard deviation in Table~\ref{table1x} refer to results over different test videos. Our EffNetV2 MS-ASCT outperforms all other methods in all considered metrics.

\begin{table}[h]
\begin{center}
\caption{Video-by-video accuracy, precision, recall, and jaccard score for different methods (mean $\pm$ std. \%)}\label{table1x}%
\resizebox{\linewidth}{!}{\begin{tabular}{lllll}
\toprule
Method name & Accuracy & Precision & Recall  & Jaccard\\
\midrule
EndoNet\cite{twinanda2016endonet}     & 81.7$\pm$4.2   & 73.7$\pm$16.1   & 79.6$\pm$7.9   &  $-$ \\
EndoNet+LSTM\cite{twinanda2017vision}     & 88.6$\pm$9.6   & 84.4$\pm$7.9   & 84.7$\pm$7.9   &  $-$ \\
MTRCNet-CL\cite{jin2020multi}     & 89.2$\pm$7.6    & 86.9$\pm$4.3   & 88.0$\pm$6.9   &  $-$ \\
\midrule
SV-RCNet\cite{jin2017sv}     & 85.3$\pm$7.3   & 80.7$\pm$7.0  & 83.5$\pm$7.5   &  $-$ \\
OHFM\cite{yi2019hard}     & 87.3$\pm$5.7   & $-$   & $-$   &  67.0$\pm$13.3 \\
TeCNO\cite{czempiel2020tecno}     & 88.6$\pm$7.8   & 86.5$\pm$7.0   & 87.6$\pm$6.7   &  75.1$\pm$6.9 \\
TMRNet(ResNeSt)\cite{jin2021temporal} & 90.1$\pm$7.6   & 90.3$\pm$3.3   & 89.5$\pm$5.0   &  79.1$\pm$5.7 \\
Trans-SVNet\cite{gao2021trans} & 90.3$\pm$7.1 & 90.7$\pm$5.0 & 88.8$\pm$7.4  &  79.3$\pm$6.6 \\
SAHC\cite{ding2022exploring} & 91.8$\pm$8.1  & 90.3$\pm$6.4  & 90.0$\pm$6.4  &   81.2$\pm$5.5 \\
EffNetV2 MS-TCN(Causal)\cite{zhang2023c}  & 93.69$\pm$5.30   & 90.69$\pm$6.44  & 91.88$\pm$5.51   &   83.06$\pm$9.51 \\
EffNetV2 Causal ASFormer\cite{zhang2023c}  & 94.48$\pm$4.22   & 92.15$\pm$5.25  & 91.50$\pm$9.45   & 84.20$\pm$10.42  \\
EffNetV2 C-ECT\cite{zhang2023c}    & 94.67$\pm$4.32   & 92.66$\pm$5.38  & 91.11$\pm$7.28   &  84.02$\pm$8.86  \\
\midrule
ResNet MS-ASCT & 94.71$\pm$4.09   & 92.71$\pm$5.31  & 91.49$\pm$6.16   &  84.35$\pm$8.69  \\
EffNetV2 MS-ASCT   & \textbf{95.26$\pm$3.49}   & \textbf{92.76$\pm$5.69}  & \textbf{92.16$\pm$9.01}  &  \textbf{85.36$\pm$10.98}\\
\bottomrule
\end{tabular}}
\end{center}
\end{table}

To further compare our methods with previous studies, we calculate the overall accuracy and segmental metrics including the segmental edit distance score, the segmental F1 score at overlapping thresholds of 10\%, 25\%, and 50\%, and their averages, as shown in Table ~\ref{tab3}. Our EffNetV2 MS-ASCT outperforms previous state-of-the-art EffNetV2 C-ECT \cite{zhang2023c} by approximately 1\% in accuracy, approximately 9\% in segmental edit distance score, and approximately 8\% in the average of the segmental F1 score at different overlapping thresholds.

\begin{table}[h]
\begin{center}
\caption{Overall accuracy (\%), segmental edit distance score, segmental F1-score at different thresholds, and the average of segmental F1-score at different thresholds \\}\label{tab3}
\resizebox{\linewidth}{!}{\begin{tabular}{lllllll}
\toprule
Method name & ACC & Edit & F1@10 & F1@25  & F1@50 & F1@AVG\\
\midrule
EffNetV2 MS-TCN(Causal) \cite{zhang2023c}  & 92.41 & 31.86 & 35.63  & 34.97 & 31.66   & 34.08\\
EffNetV2 Causal ASFormer \cite{zhang2023c}  & 93.44 & 46.20 & 55.43  & 54.60 & 50.47   & 53.50\\
EffNetV2 C-ECT \cite{zhang2023c}   & 93.53 & 53.00 & 60.61  & 59.71 & 54.10   & 58.14\\
\midrule
EffNetV2 MS-ASCT  & \textbf{94.64} & \textbf{61.82} & \textbf{68.02} & \textbf{68.02} & \textbf{62.45} & \textbf{66.16}\\  
\bottomrule
\end{tabular}}
\end{center}
\end{table}

\begin{figure}[h]%
\centering
\includegraphics[width=0.45\textwidth]{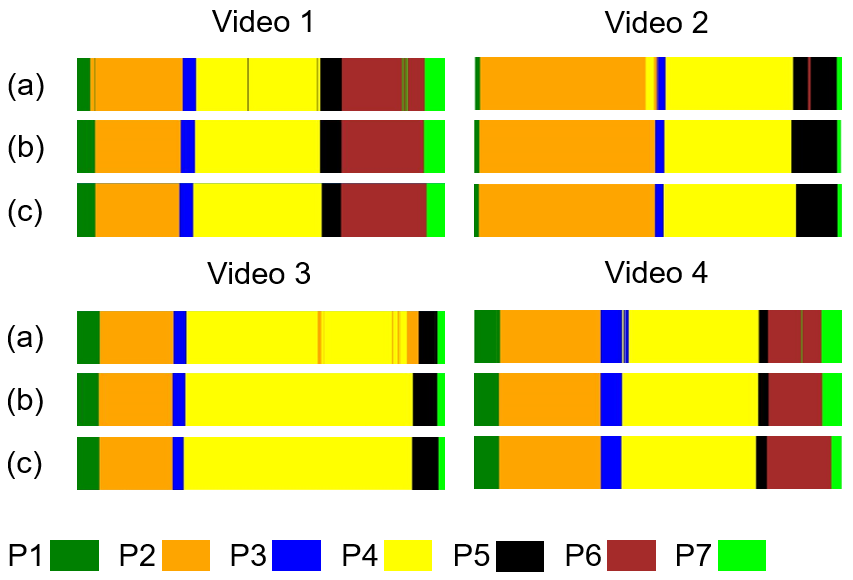}
\caption{Color-coded ribbon illustration for online surgical phase recognition: (a) EffNetV2 Causal ASFormer (b) EffNetV2 MS-ASCT (c) Ground Truth}\label{fig4}
\end{figure}

We visualize the prediction results of 4 test videos for EffNetV2 Causal ASFormer \cite{zhang2023c} and EffNetV2 MS-ASCT as shown in Figure \ref{fig4}. These visualizations demonstrate that our EffNetV2 MS-ASCT produces fewer over-segmentation errors and out-of-order predictions. 

\begin{figure}[h]%
\centering
\includegraphics[width=0.4\textwidth]{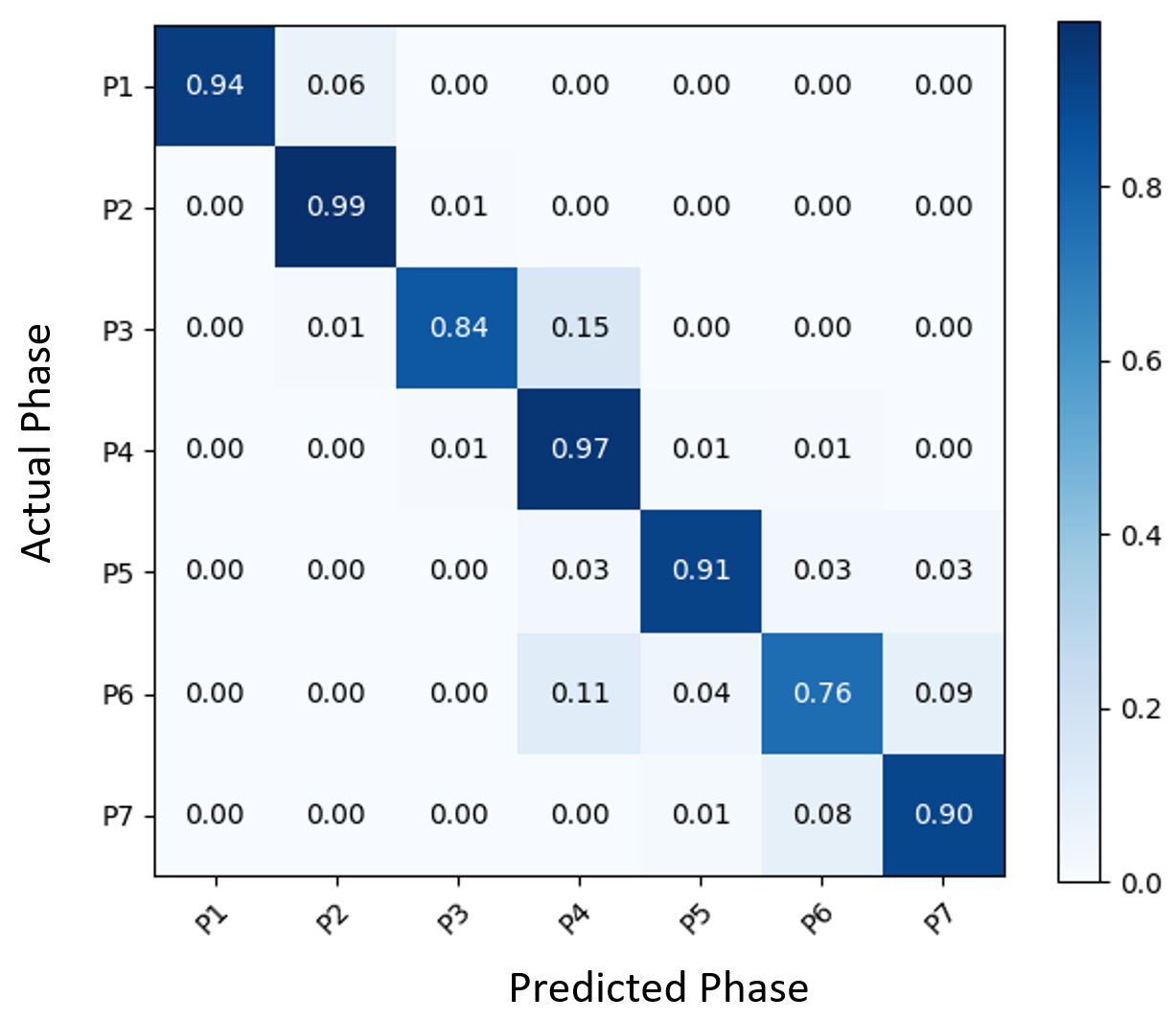}
\caption{Normalized Confusion matrix of EffNetV2 MS-ASCT on the Cholec80 dataset for online surgical phase recognition}\label{fig5}
\end{figure}

We also plot the normalized confusion matrix from one of our experiments with EffNetV2 MS-ASCT on the Cholec80 dataset for online surgical phase recognition in Figure~\ref{fig5}. Some of the prediction errors are due to our model predicting P3 as P4. Our model is also sometimes confused between P6 and P7. Both P3 and P7 are short surgical phases. Insufficient training data may be affecting the performance of these surgical phases.

\subsubsection{Offline surgical phase recognition}

To demonstrate that our ResNet MS-AST outperforms other state-of-the-art methods, we repeat our experiments multiple times and compare our ResNet MS-AST with ResNet ASFormer \cite{zhang2022surgicala}, SlowFast Transformer, SlowFast BiGRU, TimeSformer Transformer, TimeSformer BiGRU, Swin Transformer, and Swin BiGRU \cite{he2022empirical} in Table~\ref{tab1}. The mean and the standard deviation refer to results over repeated experimental runs in Table~\ref{tab1}.
Our ResNet MS-AST outperforms ResNet ASFormer\cite{zhang2022surgicala} and other methods on all considered metrics. Our EffNetV2 MS-AST achieves comparable performance with ResNet MS-AST.

\begin{table}[h]
\centering
\caption{Overall accuracy, precision, and recall for different methods on Cholec80 dataset for offline surgical phase recognition (mean $\pm$ std. \%)}\label{tab1}
\begin{tabular}{llll}
\hline
Method Name &  Accuracy & Precision & Recall\\
\hline

SlowFast Transformer \cite{he2022empirical}  & 74.12$\pm$0.03 & 69.73$\pm$0.32 & 65.99$\pm$1.15 \\
SlowFast BiGRU \cite{he2022empirical} & 92.74$\pm$0.23 & 87.71$\pm$1.27 & 84.69$\pm$1.00 \\
TimeSformer Transformer \cite{he2022empirical} & 73.46$\pm$0.99 & 73.93$\pm$0.32 & 68.27$\pm$2.28 \\
TimeSformer BiGRU \cite{he2022empirical} & 92.82$\pm$1.91 & 89.70$\pm$1.34 & 86.18$\pm$2.67 \\
Swin Transformer \cite{he2022empirical} & 80.10$\pm$0.72 & 74.35$\pm$0.98 & 74.37$\pm$0.69 \\
Swin BiGRU \cite{he2022empirical} & 93.87$\pm$0.04 & 89.96$\pm$0.79 & 89.65$\pm$0.58 \\
ResNet ASFormer \cite{zhang2022surgicala} & 93.39$\pm$0.23 & 89.85$\pm$0.42 & 89.78$\pm$0.28  \\
\hline
ResNet MS-AST & 95.17$\pm$0.07 & 90.41$\pm$0.20 & 91.85$\pm$0.13  \\
EffNetV2 MS-AST & \textbf{95.18$\pm$0.06} & \textbf{90.64$\pm$0.41} & \textbf{91.85$\pm$0.46}  \\

\hline
\end{tabular}
\end{table}

We also calculated video-by-video accuracy, precision, recall, and Jaccard score for our EffNetV2 MS-AST and compared them with other state-of-the-art approaches as shown in Table~\ref{table5xx}. The mean and the standard deviation in Table~\ref{table5xx} refer to results over different test videos. Our EffNetV2 MS-AST outperforms all other methods in all considered metrics.

\begin{table}[h]
\centering
\caption{Video-by-video accuracy, precision, recall, and jaccard for different methods on Cholec80 dataset (mean $\pm$ std. \%)}\label{table5xx}
\resizebox{\linewidth}{!}{\begin{tabular}{lllll}
\hline
Method Name &  Accuracy & Precision & Recall & Jaccard\\
\hline
Less is More(Timestamp) \cite{wang2022less} & 91.9$\pm$5.6 & 89.5$\pm$4.4 & 90.5$\pm$5.9 & 79.9$\pm$8.5 \\
Not End-to-End(TCN) \cite{yi2022not} & 92.8$\pm$5.0 & $-$ & 87.5$\pm$8.3 & 78.7$\pm$9.4 \\
ResNet MS-TCN \cite{farha2019ms,czempiel2020tecno}  & 92.88$\pm$6.15 & 92.22$\pm$4.16 & 89.76$\pm$7.23 & 82.34$\pm$6.25\\
ResNet ASFormer \cite{yi2021asformer,zhang2022surgicala} & 94.25$\pm$5.17 & 91.70$\pm$5.39 & 92.33$\pm$5.01 & 84.48$\pm$6.49 \\
\hline
ResNet MS-AST & 96.06$\pm$3.12 & 92.83$\pm$4.67 & 94.28$\pm$3.67 & 87.41$\pm$6.87 \\
EffNetV2 MS-AST & \textbf{96.15$\pm$3.94} & \textbf{93.15$\pm$5.00} & \textbf{94.57$\pm$3.52} & \textbf{88.31$\pm$6.89} \\

\hline
\end{tabular}}
\end{table}

\subsubsection{Action segmentation on non-medical datasets}
MS-TCN and ASFormer were originally designed and evaluated in the non-medical domain with datasets including the 50Salads and GTEA datasets. To demonstrate that our MS-AST can also be utilized for non-medical videos in the action segmentation domain, we evaluate our MS-AST on the 50Salads and GTEA datasets following the MS-TCN and ASFormer studies \cite{farha2019ms,yi2021asformer}. For fair comparisons and to demonstrate that our MS-AST can be trained with features generated by feature extraction networks other than image classification networks, we utilized I3D features provided by \cite{farha2019ms} as training data for our MS-AST.

As shown in Table~\ref{tab4} and Table~\ref{tab5}, our MS-AST outperforms ASFormer\cite{yi2021asformer} which utilizes only one temporal scale on all considered metrics. Our MS-AST can achieve state-of-the-art results on both datasets.

\begin{table}[h]
\begin{center}
\caption{Overall accuracy (\%), segmental edit distance score, and segmental F1-score for the 50Salads dataset \\}\label{tab4}
\resizebox{\linewidth}{!}{\begin{tabular}{llllll}
\toprule
Method name & ACC & Edit & F1@10 & F1@25  & F1@50 \\
\midrule
MS-TCN \cite{farha2019ms}  & 80.7 & 67.9 & 76.3 & 74.0 & 64.5\\
ASFormer \cite{yi2021asformer}  & 85.6 & 79.6 & 85.1 & 83.4 & 76.0\\
UVAST \cite{behrmann2022unified}   & 87.4 & 83.9 & 89.1  & 87.6 & 81.7\\
\midrule
MS-AST  & \textbf{90.5} & \textbf{85.9} & \textbf{90.2} & \textbf{89.2} & \textbf{84.7}\\  
\bottomrule
\end{tabular}}
\end{center}
\end{table}

\begin{table}[h]
\begin{center}
\caption{Overall accuracy (\%), segmental edit distance score, segmental F1-score for the GTEA dataset \\}\label{tab5}
\resizebox{\linewidth}{!}{\begin{tabular}{lllllll}
\toprule
Method name & ACC & Edit & F1@10 & F1@25  & F1@50 \\
\midrule
MS-TCN \cite{farha2019ms}  & 76.3 & 79.0 & 85.8 & 83.4 & 69.8\\
ASFormer \cite{yi2021asformer}  & 79.7 & 84.6 & 90.1 & 88.8 & 79.2\\
UVAST \cite{behrmann2022unified}   & 80.2 & \textbf{92.1} & \textbf{92.7}  & 91.3 & 81.0\\
\midrule
MS-AST  & \textbf{82.3} & 90.3 & 92.6 & \textbf{91.6} & \textbf{84.9}\\  
\bottomrule
\end{tabular}}
\end{center}
\end{table}

\section{Conclusions}
In this work, we propose Multi-Scale Action Segmentation Transformer (MS-AST) for offline surgical phase recognition and Multi-Scale Action Segmentation Causal Transformer (MS-ASCT) for online surgical phase recognition.  Our proposed network utilizes multiple scales of temporal feature modeling, which is a good conceptual fit for multi-step surgical scenarios and can work with various feature extraction networks. Our methods achieve new state-of-the-art results on the Cholec80 dataset for surgical phase recognition, representing a significant advancement in surgical phase recognition. We also test our MS-AST on the 50Salads and GTEA datasets to demonstrate the wider applicability of our method and perform similarly to or better than the state-of-the-art results in both cases.  
Future research directions could explore the integration of our models in real-time surgical environments, adapt our models to more surgical procedures, or focus on enhancing the interpretability of model predictions to provide more actionable insights for surgical teams. 

\bibliographystyle{IEEEtran}
\bibliography{conference_101719}

\end{document}